\documentclass[10pt,twocolumn,letterpaper]{article}

\usepackage{cvpr}              

\usepackage{graphicx}
\usepackage{amsmath}
\usepackage{amssymb}
\usepackage{booktabs}
\usepackage{bm}
\usepackage{color}
\usepackage{enumitem}
\usepackage{lipsum}
\usepackage[accsupp]{axessibility}
\setenumerate[1]{itemsep=0pt,partopsep=0pt,parsep=\parskip,topsep=1pt}
\setitemize[1]{itemsep=0pt,partopsep=0pt,parsep=\parskip,topsep=1pt}
\setdescription{itemsep=0pt,partopsep=0pt,parsep=\parskip,topsep=1pt}

\usepackage[pagebackref,breaklinks,colorlinks]{hyperref}

\usepackage[capitalize]{cleveref}
\crefname{section}{Sec.}{Secs.}
\Crefname{section}{Section}{Sections}
\Crefname{table}{Table}{Tables}
\crefname{table}{Tab.}{Tabs.}

\def\cvprPaperID{8685} 
\def\confName{CVPR}
\def\confYear{2023}

\begin{document}

\title{Masked and Adaptive Transformer for Exemplar Based Image Translation}



\author{Chang Jiang\textsuperscript{1},  Fei Gao\textsuperscript{1,2*}, Biao Ma\textsuperscript{1}, Yuhao Lin\textsuperscript{1}, Nannan Wang\textsuperscript{3}, Gang Xu\textsuperscript{1} \\
\textsuperscript{1} School of Computer Science and Technology, Hangzhou Dianzi University \\
\textsuperscript{2} Hangzhou Institute of Technology, Xidian University 
\textsuperscript{3} ISN State Key Laboratory, Xidian University \\
}


\maketitle

{
\let\thefootnote\relax\footnotetext{*Corresponding Author}
}

\begin{abstract}
We present a novel framework for exemplar based image translation. 
Recent advanced methods for this task mainly focus on establishing cross-domain semantic correspondence, which sequentially dominates image generation in the manner of local style control. 
Unfortunately, cross-domain semantic matching is challenging; and matching errors ultimately degrade the quality of generated images.  
To overcome this challenge, we improve the accuracy of matching on the one hand, and diminish the role of matching in image generation on the other hand. 
To achieve the former, we propose a masked and adaptive transformer (MAT) for learning accurate cross-domain correspondence, and executing context-aware feature augmentation. 
To achieve the latter, we use source features of the input and global style codes of the exemplar, as supplementary information, for decoding an image. 
Besides, we devise a novel contrastive style learning method, for acquire quality-discriminative style representations, which in turn benefit high-quality image generation.
Experimental results show that our method, dubbed MATEBIT, performs considerably better than state-of-the-art methods, in diverse image translation tasks.  
The codes are available at \url{https://github.com/AiArt-HDU/MATEBIT}.
\end{abstract}

\section{Introduction}
\label{sec:intro}

Image-to-image translation aims at transfer images in a source domain to a target domain \cite{isola2017image, zhu2017unpaired}. Early studies learn mappings directly by Generating Adversarial Networks (GANs), and have shown great success in various applications \cite{wang2018esrgan, chang2018pairedcyclegan}. 
Recently, exemplar based image translation \cite{ma2018exemplar,zhang2020cocosnet,luo2022memory}, where an exemplar image is used to control the style of translated images, has attracted a lot of attention. 
Such methods allow high flexibility and controllability, and have a wide range of potential applications in social networks and metaverse. For example, people can transfer a facial sketch to an artistic portrait, in the style of oil paintings or avatars. Despite the remarkable progress, yielding high-fidelity images with consistent semantic and faithful styles remains a grand challenge.

Early pioneering works \cite{kipf2016variational, park2019SPADE,huang2018multimodal} attempt to globally control the style of generated images. However, such methods ignore spatial correlations between an input image and an exemplar, and may fail to produce faithful details.
Recently, some advanced methods \cite{zhang2020cocosnet, zhou2021cocosnetv2, zhan2022marginal, DynaST} first establish the cross-domain semantic correspondence between an input image and an exemplar, and then use it to warp the exemplar for controlling local style patterns. 
In these methods, the quality of generated images relies heavily on the learned correspondence \cite{seo2022midms}. 
Unfortunately, cross-domain semantic matching is challenging, since there is no reliable supervision on correspondence learning \cite{zhang2020cocosnet}. As a result, potential matching errors ultimately lead to degraded artifacts in generated images.

\begin{figure}
	\centering
	\includegraphics[width=1\linewidth]{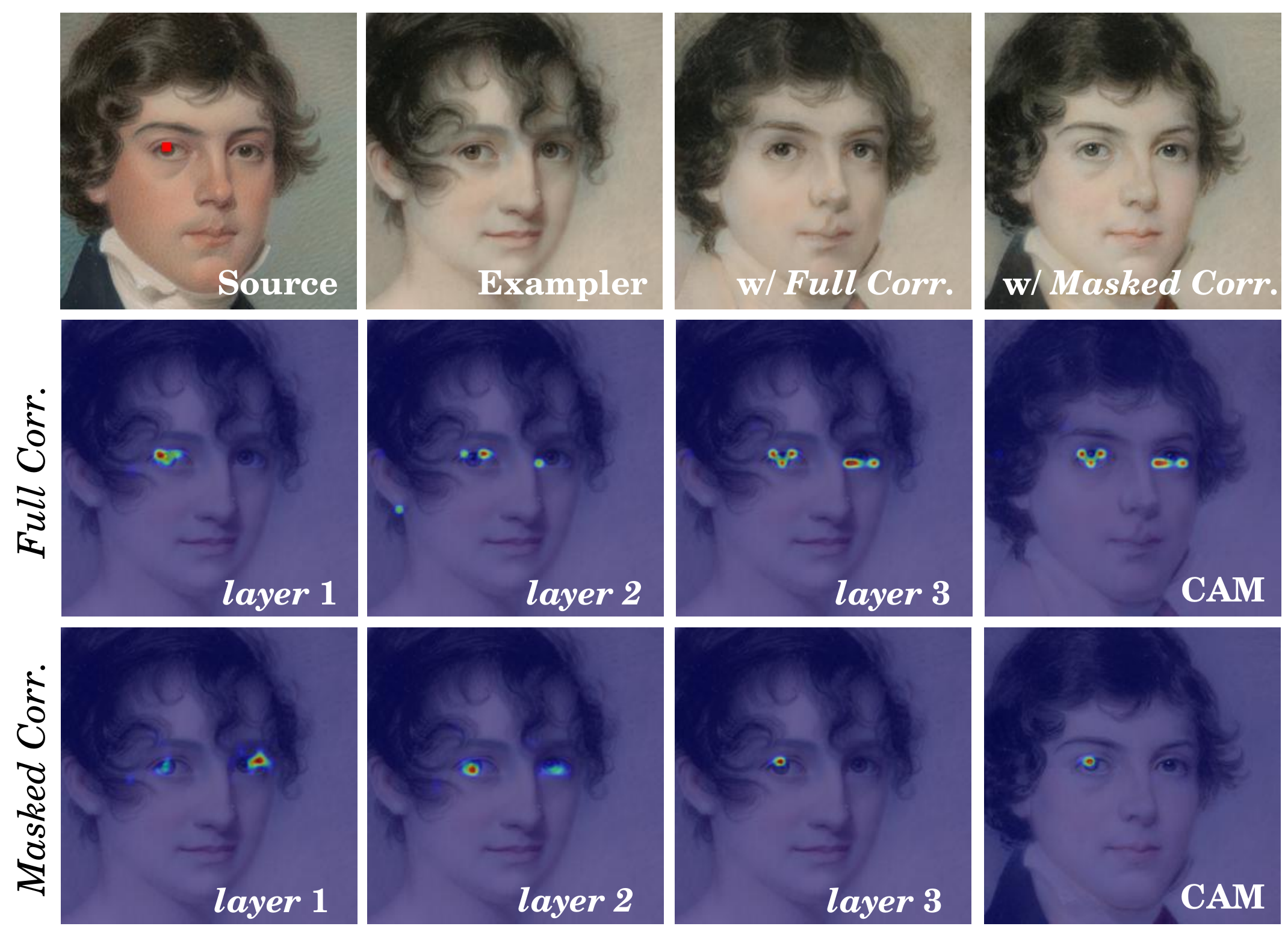}
		 \vspace{-0.6cm}
	\caption{Visualization of correspondence maps. The {\color{red} red} point is the query position. \textit{Full Corr.} and \textit{Masked Corr.} denote the full correspondence \cite{zhang2020cocosnet} and masked one in our method, respectively. CAM denotes visualization by \textit{Class Activation Mapping} \cite{zhou2016CAM}.}
	\label{fig:visatt}
	 \vspace{-0.5cm}
\end{figure}

To combat this challenge, we propose to boost the matching accuracy on one hand, and to diminish the role of matching in image generation on the other hand. 
Inspired by the great success of Transformers  \cite{vaswani2017attention,dosovitskiy2020image,liu2021swin,heo2021rethinking}, we first devise a \textit{Masked and Adaptive Transformer} (MAT) for learning accurate cross-domain correspondence and executing context-aware feature augmentation. 
Previous works \cite{zhang2020cocosnet, zhou2021cocosnetv2, zhan2022marginal} have used the vanilla attention mechanism \cite{vaswani2017attention} for learning full correspondence. 
However, the initial attention typically involves ambiguous correspondences (2nd row in Fig. \ref{fig:visatt}). 
To mitigate these limitations, in MAT, we use a masked attention to distinguish the correspondence as reliable or not, and then reliability-adaptively aggregate representations. 
Besides, the \textit{Feed-Forward Network} (FFN) \cite{vaswani2017attention} in vanilla transformers neglects contextual correlations inside an image. 
We thus replace FFN by an adaptive convolution block \cite{liu2022convnext}, where the coordinate attention \cite{hou2021coordinate} and depthwise separable convolution \cite{chollet2017xception} are used to capture contextual correlations and to improve efficiency. With a joint consideration of matching reliability and contextual correlations, MAT gradually focuses on accurate correspondences and emphasizes on features of interest (3rd row in Fig. \ref{fig:visatt}).

In addition, to boost both the semantic consistency and style faithfulness, we supplementally use semantic features of the input image and global style codes of the exemplar for decoding an image. 
To this end, we first design our whole network following the U-Net architecture \cite{isola2017image}. 
Besides, we devise a novel contrastive style learning (CSL) framework for acquiring discriminative style representations.
Recently, Zhang et al. \cite{zhang2020cast} propose a similar CSL method, where the target exemplar is used as a positive sample, and the other exemplars as negative ones. 
Differently, we use low-quality images, generated during early training stages, as negative samples. In this way, our style codes are desired to discriminate not only subtle differences in style, but also those in perceptual quality. 
Ultimately, the learned \textit{global} style codes, cooperating with the \textit{local} style control induced by MAT, in turn benefit high-quality image generation. 

With the proposed techniques above, our full model, dubbed MATEBIT, diminishes the impact of position-wise matching on image quality, and integrates both local and global style control for image generation. 
Experimental results show that MATEBIT generates considerably more plausible images than previous state-of-the-art methods, in diverse image translation tasks.
In addition, comprehensive ablation studies demonstrate the effectiveness of our proposed components. 
Finally, we perform interesting applications of photo-to-painting translation and Chinese ink paintings generation. 


\section{Relate Work}
\label{sec:related}

\textbf{Exemplar Based Image Translation.} 
Recently, exemplar based image translation has attracted increasing attention. For example, Park et al. \cite{park2019SPADE} learn an encoder to map the exemplar image into a global style vector, and use it to guide image generation. Such a global style control strategy enables style consistency in whole, but fails to produce subtle details. 
Most recently, researchers propose a matching-then-generation framework \cite{seo2022midms}. Specially, they first establish dense correspondence between an input and an exemplar, and then reshuffle the exemplar for locally control the style of synthesize images. For example, Zhang et al.\cite{zhang2020cocosnet} establish position-wise correspondence based on the Cosine attention mechanism and warp the exemplar correspondingly. Afterwards, the warped image dominates the generation of images in the manner of SPADE \cite{park2019SPADE}. To reduce the cost of matching in high-resolution image generation, Zhou et al.\cite{zhou2021cocosnetv2} introduce a hierarchical refinement of semantic correspondence from ConvGRU-PatchMatch. Besides, Liu et al.\cite{DynaST} used a dynamic pruning method for learning hierarchical sparse correspondence. They also use reliability-adaptive feature integration to improve the quality of generated images. 

Previous methods merely use global or local style control, and the latter relies heavily on the learned correspondence. Besides, they consider little about contextual correlations inside an image. 
In this paper, we use both global and local style control to boost the style consistency. Besides, we take contextual correlations into consideration and execute reliability-adaptive feature augmentation. 

\textbf{Transformers.}
Transformers \cite{vaswani2017attention} have shown incredible success from the field of natural language processing (NLP) \cite{kenton2019bert} to computer vision (CV) \cite{dosovitskiy2020image,liu2021swin}. 
Multi-head attention (MHA) and FFN are key components in a Transformer, and have been used in exemplar based image translation. However, they induce unreliable matching results and neglect context correlations in feature translation. In our MAT, we combat these limitations by replacing them with a masked attention and a context-aware convolution block, respectively. 
Recently, researchers use semantic masks to facilitate representation learning \cite{rao2021dynamicvit,he2022MAE,cheng2022masked}, where a mask predictor is required. 
Differently, we use a ReLU function to mask over the attention layer, for distinguishing correspondence as reliable or not (Sec. \ref{ssec:MAT}). In general, MAT follows a concise and efficient architecture. 


\textbf{Contrastive Learning.}
Contrastive learning has shown its effectiveness in various computer vision tasks \cite{he2020momentum,park2020contrastive,hu2022qs}. 
The basic idea is to learn a representation by pushing positive samples toward an anchor, and moving negative samples away from it. Different sampling strategies and contrastive losses have been extensively explored in various downstream tasks. For example, Chen et al.\cite{chen2020simple} and He et al.\cite{he2020momentum} obtain positive samples by augmenting original data. In the field of image translation, Park et al.\cite{park2020contrastive} propose patch-wise contrastive learning by maximizing the mutual information between cross-domain patches. Similarly, Zhang et al. \cite{zhang2020cast} use contrastive learning for acquiring discriminative style representations. In the task of exemplar based image translation, Zhan et al.\cite{zhan2022marginal} use contrastive learning to align cross-domain images to a consistent semantic feature space, so as to boost the accuracy of matching. 
Differently, we use early generated images as negative samples, so that the learned style representations can discriminate subtle differences in both style and perceptual quality (Sec. \ref{ssec:StyleCL}). 

\section{The Proposed Method}
\label{sec:method}

\begin{figure}
	\centering
	\includegraphics[width=1.0\linewidth]{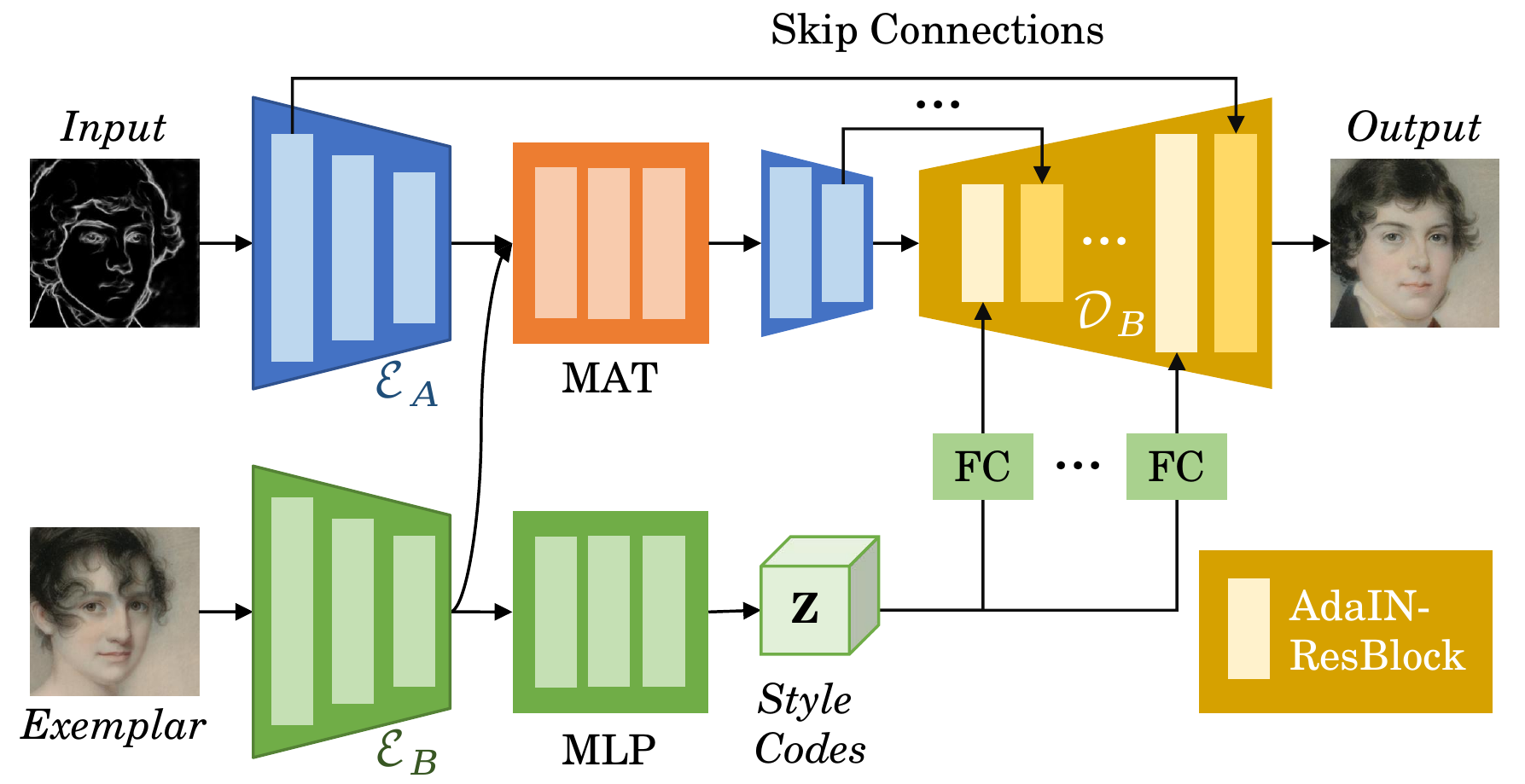}
	 \vspace{-0.6cm}
	\caption{Overview of our image translation network, MATEBIT.
	}
	\label{fig:main}
	 \vspace{-0.5cm}
\end{figure}

Given an input image $x_A$ in domain $\mathcal{A}$ and an exemplar image $y_B$ in domain $\mathcal{B}$, our goal is to generate a target image $x_B$ which preserves semantic structures in $x_A$ but resembles the style of similar parts in $y_B$.   
Fig. \ref{fig:main} shows an overview of our translation network $\mathcal{G}$. 
Specially, we first align $x_A$ and $y_B$ to an intermediate feature space by encoders $\mathcal{E}_A$ and $\mathcal{E}_B$, respectively. 
Afterwards, we use a \textit{Masked and Adaptive Transformer} (MAT) for correspondence learning and feature augmentation. Finally, a decoder $\mathcal{D}_B$ produces an output image $\hat{x}_B$ based on the augmented features, as well as the source features and target style codes. Details are described below.

\subsection{Masked and Adaptive Transformer (MAT)}
\label{ssec:MAT}

\begin{figure}
	\centering
	\includegraphics[width=0.9\linewidth]{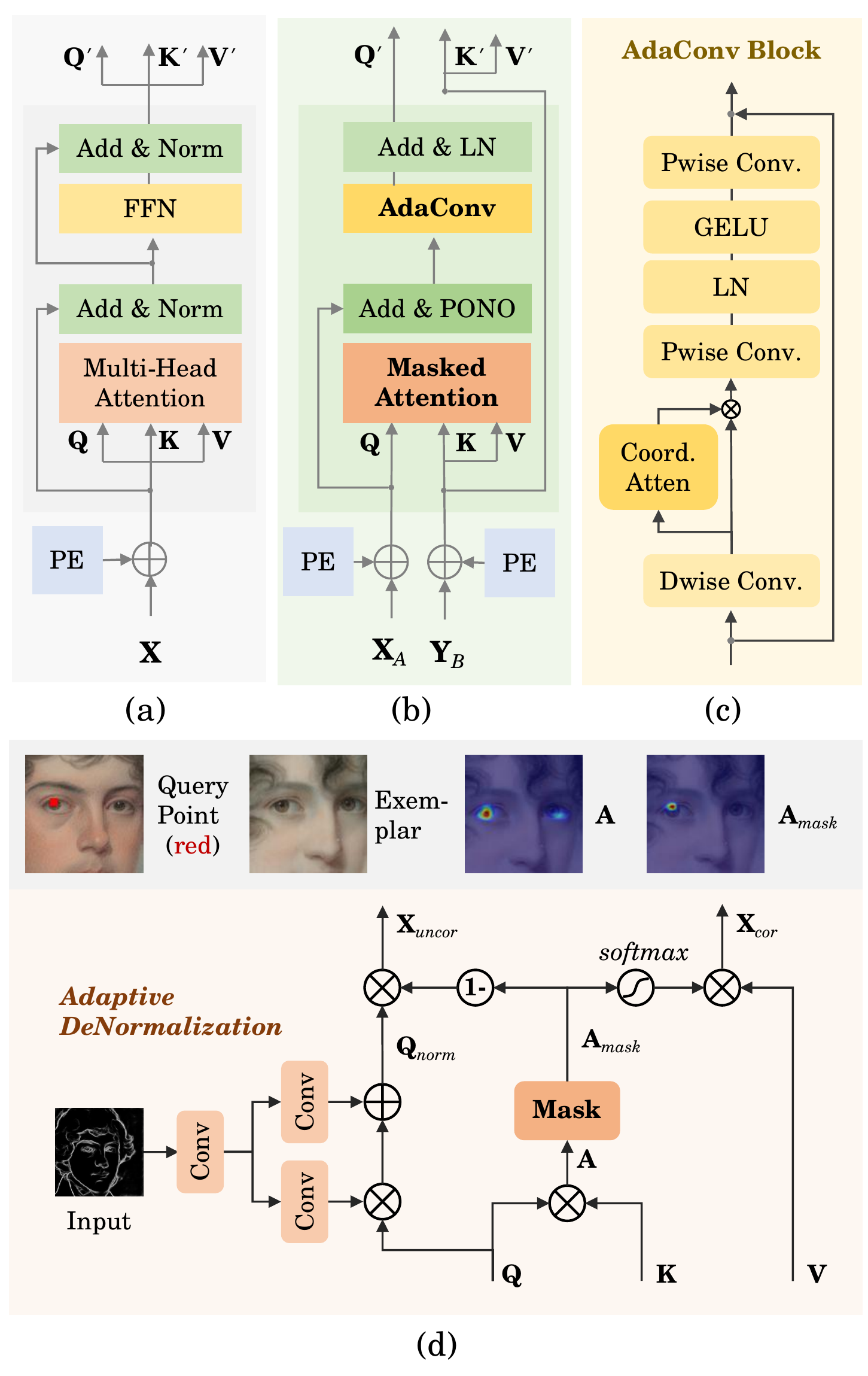}
	 \vspace{-0.5cm}
	\caption{Detailed architectures. (a) Vanilla Transformer block, (b) MAT block, (c) AdaConv block, and (d) \textit{Masked Attention}. 
	}
	\label{fig:mat}
	 \vspace{-0.4cm}
\end{figure}

In order to establish accurate cross-domain correspondence, we propose a novel and concise Transformer architecture, i.e. MAT. 
In general, the  architecture of MAT (Fig. \ref{fig:mat}b) follows that of vanilla Transformers (Fig. \ref{fig:mat}a) \cite{vaswani2017attention}. 
Differently, we use masked attention to distinguish reliable and unreliable correspondence, instead of using multi-head attention. 
Besides, we use \textit{Positional Normalization} (PONO) \cite{li2019positional} and an \textit{Adaptive Convolution} (AdaConv) block \cite{liu2022convnext}, instead of LN and MLP-based FFN, respectively. 
MAT is desired to gradually concentrate on accurate matching, and to reliability-adaptively augment representations with contextual correlations.

\textbf{Masked Correspondence Learning.}
Let $\mathbf{X}_A \in \mathbb{R}^{H \times W \times C}$ and $\mathbf{Y}_B \in \mathbb{R}^{H \times W \times C}$ be the representations of $x_A$ and $y_B$ in the intermediate feature space, with height $H$, width $W$, and $C$ channels.
We first map $\mathbf{X}_A$ to the query $\mathbf{Q} \in \mathbb{R}^{HW \times C}$, and $\mathbf{Y}_B$ to the key $\mathbf{K} \in \mathbb{R}^{HW \times C}$ and value $\mathbf{V} \in \mathbb{R}^{HW \times C}$, by using $1 \times 1$ convolutions, respectively. As shown in Fig. \ref{fig:mat}d, we add positional encoding (PE) to $\mathbf{X}_A$ and $\mathbf{Y}_B$, for embedding spatial correlations.
Afterwards, we learn the initial correspondence $\mathbf{A} \in \mathbb{R}^{HW \times HW}$ following the Cosine attention mechanism \cite{zhang2020cocosnet}, i.e.
\begin{equation}
		\mathbf{A}(u,v) = \frac{\tilde{\mathbf{Q}}(u) \tilde{\mathbf{K}}(v)^{T} }{|| \tilde{\mathbf{Q}}(u) || \cdot || \tilde{\mathbf{K}}(v) ||}, 
	\label{eq:atten}
\end{equation}
with 
$
		\tilde{\mathbf{Q}}(u) = \mathbf{Q}(u) - \bar{\mathbf{Q}}(u), ~
		 \tilde{\mathbf{K}}(v) = \mathbf{K}(v) - \bar{\mathbf{K}}(v),
$
where $u, v \in [1, ..., HW]$ are position indices; $\bar{\mathbf{Q}}(u)$ and $\bar{\mathbf{K}}(v)$ are the means of $\mathbf{Q}(u)$ and $\mathbf{K}(v)$, respectively. $\mathbf{A}(u,v)$ is the matching score between $\mathbf{Q}(u)$ and $\mathbf{K}(v)$.

Previous methods \cite{zhang2020cocosnet, zhan2022marginal} typically use the initial correspondence map $\mathbf{A}$ to reshuffle an exemplar for controlling local patterns in image synthesis. However, induced by the difficulties in cross-domain correspondence learning, $\mathbf{A}$ involves unreliable match scores (Fig. \ref{fig:mat}d). As a result, the reshuffled image will lead to implausible artifacts in generated images. 
To combat this limitation, we distinguish initial matching scores as reliable or not, according to their signs \cite{min2019hyperpixel}. 
The masked correspondence map becomes: 
\begin{equation}
	\mathbf{A}_{mask} = \mathrm{ReLU}(\mathbf{A}) ,
	\label{eq:mattn}
\end{equation}
In DynaST \cite{DynaST}, two networks are used to predict the reliability mask of correspondence. 
However, it's challenging to effectively train the network, because there is no supervision on matching during training. In contrast, ReLU contains no learnable parameters and ultimately leads to superior performance over DynaST (Sec. \ref{ssec:comp}). 


\textbf{Reliability-Adaptive Feature Aggregation.} 
For regions with reliable correspondence in $x_A$, we use  $\mathbf{A}_{mask}$ to warp the value features, $\mathbf{V}$, derived from the exemplar:
\begin{equation}
	\mathbf{X}_{cor}  = \tilde{\mathbf{A}}_{mask} \mathbf{V}, 
	\text{~with~} \tilde{\mathbf{A}}_{mask} = \mathrm{softmax}(\alpha \cdot \mathbf{A}_{mask}),
	\label{eq:xcor}
\end{equation}
where $\alpha$ is a scaling coefficient to control the sharpness of the softmax function. In default, we set its value as $100$. 

For regions with unreliable correspondence in $x_A$, $\mathbf{X}_{cor}$ provides an average style representation of $\mathbf{V}$. 
We further extract complementary information from the query, $\mathbf{Q}$, derived from the input. 
Inspired by SPADE \cite{park2019SPADE}, we first transfer $\mathbf{Q}$ to the target domain by using pixel-wise modulation parameters (i.e., $\boldsymbol\gamma$ for scale and $\boldsymbol\beta$ for bias) learned from $x_A$.  
The modulation is formulated by:
\begin{equation}
		\mathbf{Q}_{norm} = \bm{\gamma}(x_A)  \frac{\mathbf{Q}-\mu({\mathbf{Q})}}{\sigma(\mathbf{Q})} + \bm{\beta}(x_A), 
		\label{eq:spade}
\end{equation}
where $\mu(\mathbf{Q})$ and $\sigma(\mathbf{Q})$ are the mean value and standard deviance of $\mathbf{Q}$. 
Afterwards, we select the translated features of unreliably corresponded regions in $x_A$ by: 
\begin{equation}
		\mathbf{X}_{uncor} = (1 - \sum\nolimits_{j}\mathbf{A}_{mask}) \odot \mathbf{Q}_{norm}, 
		\label{eq:xuncor}
\end{equation}
where the summation is along the second dimension; $\odot$ denotes point-wise production with broadcasting. Since $\boldsymbol\gamma$ and $\boldsymbol\beta$ are learned from the input image $x_A$, the modulated features preserve the semantic information of $x_A$. Besides, constraints on the generated image will push the selected features convey to the style of $y_B$. 

Ideally, $\mathbf{X}_{cor}$ and $\mathbf{X}_{uncor}$ would complement each other and facilitate both semantic consistency and style relevance in image generation. 
To this end, we integrate $\mathbf{X}_{cor}$, $\mathbf{X}_{uncor}$, and $\mathbf{Q}$ by: 
\begin{equation}
	\mathbf{X}_{agg} = \mathrm{PONO}(\mathbf{X}_{cor} + \mathbf{X}_{uncor} + \mathbf{Q}).
	\label{eq:xfuse}
\end{equation}
In PONO \cite{li2019positional}, features at each position are normalized dependently. Compared to LN in vanilla transformers and DynaST \cite{DynaST}, PONO boosts the flexibility in reliability-adaptive feature aggregation.


\textbf{Context-Aware Feature Augmentation.} 
Inspired by ConvNeXT \cite{liu2022convnext}, we replace FFN by an AdaConv block to position-adaptively emphasize informative representations.
Besides, we use the \textit{coordinate attention} (CoordAtten) module \cite{hou2021coordinate} to capture contextual correlations. 

The architecture of the AdaConv block is as shown in Fig. \ref{fig:mat}c. 
We fist use the depthwise convolution (Dwise) to update representations in each channel separately; and then use two pointwise convolutions (Pwise) to automatically emphasize representations of interest, at every position. The \textit{Gaussian Error Linear Unit} (GELU) activation function and \textit{Layer Norm} (LN) are used after the first Pwise layer \cite{liu2022convnext}. 
Notably, CoordAtten is used after the Dwise layer for modeling long-range dependencies in an image. Specially, CoordAtten produces cross-channel and position-sensitive attention maps, which helps our model to more accurately locate the representations of interest \cite{hou2021coordinate}.  

Finally, the output of a MAT block is obtained with a residual connection, i.e.
$
\mathbf{X}_{\mathrm{MAT}} = \mathrm{AdaConv}(\mathbf{X}_{agg}) + \mathbf{X}_{agg}. 
$
In the implementation, we stack three MAT blocks in default to gradually refine the correspondence and to augment informative representations (Fig. \ref{fig:visatt}). Empirical verifications will be given in Sec. \ref{ssec:ablation}.

\textbf{Benefits of MAT.} Fig. \ref{fig:mat}d illustrates the impact of MAT. The query point locates over the left eye of the source image. Here we show the magnitudes of its correspondence over the exemplar, in the third layer of MAT. Obviously, the original correspondence $\mathbf{A}$ covers both eyes of the exemplar. In contrast, the masked correspondence $\mathbf{A}_{mask}$ accurately concentrates over the left eye. 
Such superiority significantly boost the quality of ultimate images. 


\subsection{Contrastive Style Learning (CSL)}
\label{ssec:StyleCL}

\begin{figure}
	\centering
	\includegraphics[width=1\linewidth]{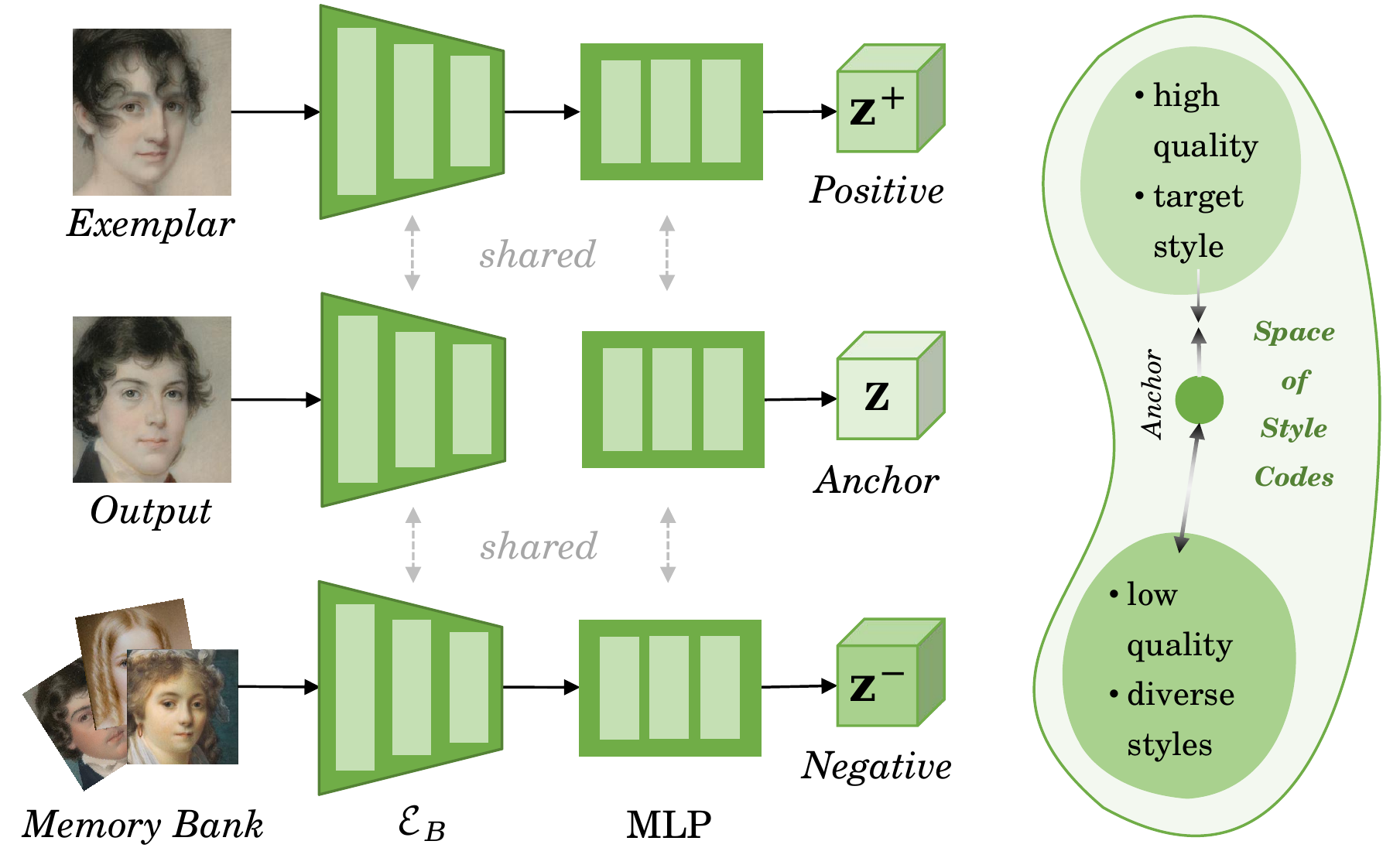}
	\caption{Contrastive style learning. The memory bank consists of divergent low-quality images generated in early training stages.}
	\label{fig:style}
	 \vspace{-0.4cm}
\end{figure}

In MATEBIT, we use the encoder $\mathcal{E}_B$ to extract local style information $\mathbf{X}_B$, and then a MLP to extract global style codes $\mathbf{z}$. $\mathbf{X}_B$ and $\mathbf{z}$ perform local and global style control on generated images, respectively (Sec. \ref{ssec:TransNet}). 
To boost the discriminative capacity of style representations, as well as the quality of generated images, we propose a novel \textit{contrastive style learning} (CSL) method (as shown in Fig \ref{fig:style}).

In our settings, the exemplars are drawn by human artists and thus considered as high-quality. In contrast, the images generated in early training stages are typically low-quality. Inspired by the idea of contrastive learning \cite{he2020momentum}, we use the exemplar $y_B$ as the positive sample, while a collection of early generated images as negative. 
Let $\mathbf{z}$ denotes style codes of the generated image $\hat{x}_B$, $\mathbf{z}^+$ that of exemplar $y_B$, and $\{ \mathbf{z}^-_{1}, \mathbf{z}^-_{2}, ..., \mathbf{z}^-_{m} \}$ the style codes of $m$ negative samples.
CSL learns style representations by maximizing the mutual information between anchors and positive samples, while minimizing that between anchors and negative samples. 
Our contrastive style loss is computed by:
\begin{equation}
\mathcal{L}_{style} = - \log \frac{ \exp(\frac{\mathbf{z}^T \mathbf{z}^{+}}\tau) }{ \exp(\frac{\mathbf{z}^T \mathbf{z}^{+}} \tau) + \sum _{j=1}^{m}\exp(\frac{\mathbf{z}^T \mathbf{z}^{-}_j}\tau)},
\label{eq:csl}
\end{equation}
where $\tau  = 0.07$ and $m = 1024$. In the implementation, we use a queue to cache negative style vectors.

\subsection{Translation network}
\label{ssec:TransNet}
To boost both the semantic consistency and style faithfulness, we additionally use source semantic features and global style codes for decoding an image. 
Specially, we design our whole translation network following U-Net (Fig. \ref{fig:main}), where the multi-level features in $\mathcal{E}_A$ are skip-connected to the decoder $\mathcal{D}_B$, for supplementing informative sematic structures of the input image $x_A$. 
Besides, we use the style codes $\mathbf{z}$ to globally control the style of generated images, in the manner of AdaIN \cite{huang2017adain}. Specially, $\mathbf{z}$ is mapped to channel-wise modulating factors by fully-connected (FC) layers. In this way, we diminish the impact of correspondence learning on image generation, and provide reliable style control for even unmatched regions.

In summary, our translation network allows both local and global style control, and reuses the semantic features of input images. As a result, the generated image is desired to present consistent semantic to the input $x_A$ and faithful style to the exemplar $y_B$. More details of our network are available in the supplementary material.



\subsection{Loss functions}
\label{ssec:Loss}

Our whole network is end-to-end optimized to jointly achieve high-fidelity image generation and accurate correspondence. 
Following \cite{zhang2020cocosnet}, we obtain training triplets $\{x_A, y_B, x_B\}$ from the ready-made data pair $\{x_A, x_B\}$, where $y_B$ is a geometrically warped version of $x_B$. The generated image is denoted by $\hat{x}_B = \mathcal{G}(x_A, y_B)$.
Our loss functions are similar to \cite{zhang2020cocosnet}, except for the previous contrastive style loss $\mathcal{L}_{style}$ and the structural loss $\mathcal{L}_{str}$ below.

\textbf{Semantic alignment Loss.} For accurate cross-domain correspondence learning, the encoders $\mathcal{E}_A$ and $\mathcal{E}_B$ should align $x_A$ and $x_B$ to consistent representations. The corresponding semantic alignment loss is:
\begin{equation}
\mathcal{L}_{align} = \left \| \mathcal{E}_A(x_A) - \mathcal{E}_B(x_B) \right \|_{1}.
	\label{eq:lossalign}
\end{equation}

\textbf{Correspondence Loss.} Ideally, if we warp $y_B$ in the same way as Eq.\ref{eq:xcor}, the resulting image should be exactly $x_B$. We thus constrain the learned correspondence by:
\begin{equation}
	\mathcal{L}_{corr} = \left \| \tilde{\mathbf{A}}_{mask}^T y_B \downarrow - {x_B} \downarrow \right \|_1,
	\label{eq:losscorr}
\end{equation}
where $\downarrow$ indicates down-sampling $y_B$ and $x_B$ to the size (i.e. width and height) of $\mathbf{X}_A$.


\textbf{Perceptual Loss.} The generated image $ \hat{x}_B $ should be semantic-consistent with the ground truth $x_B$ in term of semantic. We thus use the perceptual loss:
\begin{equation}
	\mathcal{L}_{perc}=\left \| \varphi_l (\hat{x}_B)- \varphi_l(x_B)\right \|_{1},
	\label{eq:lossperc}
\end{equation}
where $\varphi_l$ denotes the activations after layer $relu4\_2 $ in pretrained VGG19 \cite{simonyan2014very} , which represent high-level semantics.

\textbf{Contextual Loss.} In addition, the generated image should be in the same style as the exemplar. In addition to the previous contrastive style loss (Eq.\ref{eq:csl}), we additionally use the contextual loss (CX) \cite{mechrez2018contextual} to constrain on local style consistency. The contextual loss is computed by:
\begin{equation}
	\mathcal{L}_{ctx} = -\log \left(\sum_l w_l \mathrm{CX}( \varphi_l(\hat{x}_B), \varphi_l(y_B) ) \right)
	\label{eq:lossctx}
\end{equation}
where $w_l$ balances the terms of different VGG19 layers.

\textbf{Structural Loss.} The generated image should preserve semantic structures in the input image. Correspondingly, we use the \textit{Learned Perceptual Image Patch Similarity} (LPIPS)  \cite{zhang2018unreasonable} between their boundaries as the structural loss:
\begin{equation}
	\mathcal{L}_{str} = \mathrm{LPIPS}(\mathcal{H}(\hat{x}_B), \mathcal{H}(x_B)),
	\label{eq:lossid}
\end{equation}
where $\mathcal{H}$ is the HED algorithm \cite{xie2015holistically}, which has been widely used for extracting semantic boundaries in an image.


\begin{table*}
	\begin{center}
		\caption{Comparison on the Metfaces \cite{karras2020training}, CelebA-HQ \cite{lee2020maskgan}, Ukiyo-e \cite{pinkney2020ukiyoe},Cartoon \cite{pinkney2020ukiyoe}, AAHQ \cite{liu2021blendgan}, and DeepFashion \cite{DeepFashion} datasets.}
		 \vspace{-0.3cm}
		\label{tab:comp}
		\renewcommand\arraystretch{1.2}
		\resizebox{1.0\textwidth}{!}{
			\begin{tabular}{l|ccccc|cc|cc|cc|cc|cc|c}				
				\toprule
				& \multicolumn{5}{c|}{CelebA-HQ} & \multicolumn{2}{c|}{Metfaces}& \multicolumn{2}{c|}{Cartoon} & \multicolumn{2}{c|}{Ukiyo-e}& \multicolumn{2}{c|}{AAHQ}& \multicolumn{2}{c|}{DeepFashion} & Time $\downarrow$ \\
				\cmidrule(lr){2-6}  \cmidrule(lr){7-8}  \cmidrule(lr){9-10}  \cmidrule(lr){11-12} 	\cmidrule(lr){13-14} \cmidrule(lr){15-16} 			
				& FID $\downarrow$ & SWD $\downarrow$ & Texture $\uparrow $ & Color $\uparrow $ & Semantic$\uparrow $&  FID $\downarrow$ & SWD $\downarrow$ & FID $\downarrow$ & SWD $\downarrow$ & FID $\downarrow$ & SWD $\downarrow$ & FID $\downarrow$ & SWD $\downarrow$  & FID $\downarrow$ & SWD $\downarrow$& ($s$) \\
				\midrule
				SPADE \cite{park2019SPADE} & 31.5 & 26.9 &  0.927 &  0.955  &  0.922   &  45.6 &  26.9 &  97.5  &  30.5  & 45.6 & 26.9 & 79.4 & 32.1 &36.2 & 27.8& \underline{0.196}  \\
				CoCosNet \cite{zhang2020cocosnet} & 14.3 & 15.2 &  0.958 &  0.977  &  0.949 &  25.6 &  24.3 &  66.8 & 27.1 & 38.3 & 13.9 & 62.6 & \underline{21.9} & 14.4&17.2 &  0.321 \\ 
				CoCosNet-v2 \cite{zhou2021cocosnetv2} & 13.2 & 14.0 & 0.954 & 0.975  & 0.948   &  \textbf{23.3} &  \underline{22.4} & 66.4 & \underline{27.0} & \underline{32.1} & \textbf{11.0} & 62.4 & 22.8 &13.0 & 16.7& 1.573 \\
				MCL-Net \cite{zhan2022marginal} & 12.8 & 14.2 &  0.951 &  0.976 & \textbf{0.953}  &  \underline{23.8} &  24.5 & 67.9 & 27.9 & 32.4 & 12.4 & \underline{64.4} & 22.2 & 12.9& 16.2& 0.309\\
				DynaST \cite{DynaST} & \underline{12.0} & \textbf{12.4} &  \underline{0.959} &  \underline{0.978}  & \underline{0.952}   & 29.2 & 28.6  & \textbf{62.8} & \textbf{26.5} & 38.9 & 14.2 & 67.2 & 24.0 & \underline{8.4}& \underline{11.8}& 0.214\\
				MATEBIT (ours) & \textbf{11.5}	 & \underline{13.2}	 & \textbf{0.966}	 & \textbf{0.986}	 & 0.949   & 26.0 & \textbf{19.1} & 	\underline{64.4} & 	27.6 & 	\textbf{30.3} & \underline{11.5} & \textbf{56.0} & \textbf{19.5} & \textbf{8.2}&\textbf{10.0} & \textbf{0.185}\\
				\bottomrule
			\end{tabular}
		}
	\end{center}
	 \vspace{-0.7cm}
\end{table*}

\textbf{Adversarial loss.} Finally, we add a discriminator $\mathcal{D}$ to distinguish real images in domain $\mathcal{B}$ and the generated images \cite{goodfellow2020generative}. The adversarial loss is:
\begin{equation}
	\begin{split}
		& \mathcal{L}_{adv}^{\mathcal{D}}= -\mathbb{E}[h(\mathcal{D}(y_B))] - \mathbb{E}[h(-\mathcal{D}(\hat{x}_B))],\\
		& \mathcal{L}_{adv}^{\mathcal{G}}= -\mathbb{E}[\mathcal{D}(\hat{x}_B)],
	\end{split}
\end{equation}
where $ h(t) = \min(0, -1 + t) $ is the hinge loss function \cite{brock2018large}.

\textbf{Total loss.} In summary, our overall objective function is,
\begin{equation}
	\begin{split}
		\underset{\mathcal{G}}{\min} ~\underset{\mathcal{D}}{\max}~ 
		& \lambda_1\mathcal{L}_{style} + \lambda_2\mathcal{L}_{align} + \lambda_3 \mathcal{L}_{corr}  + \lambda_4 \mathcal{L}_{str} \\
		& + \lambda_5 (\mathcal{L}_{perc} + \mathcal{L}_{ctx}) + \lambda_6(\mathcal{L}_{adv}^\mathcal{G}	+ \mathcal{L}_{adv}^{\mathcal{D}})
	\end{split}
\end{equation}
where $\lambda$ denotes the weight parameters.

\begin{figure*}
	\centering
	\includegraphics[width=1\linewidth]{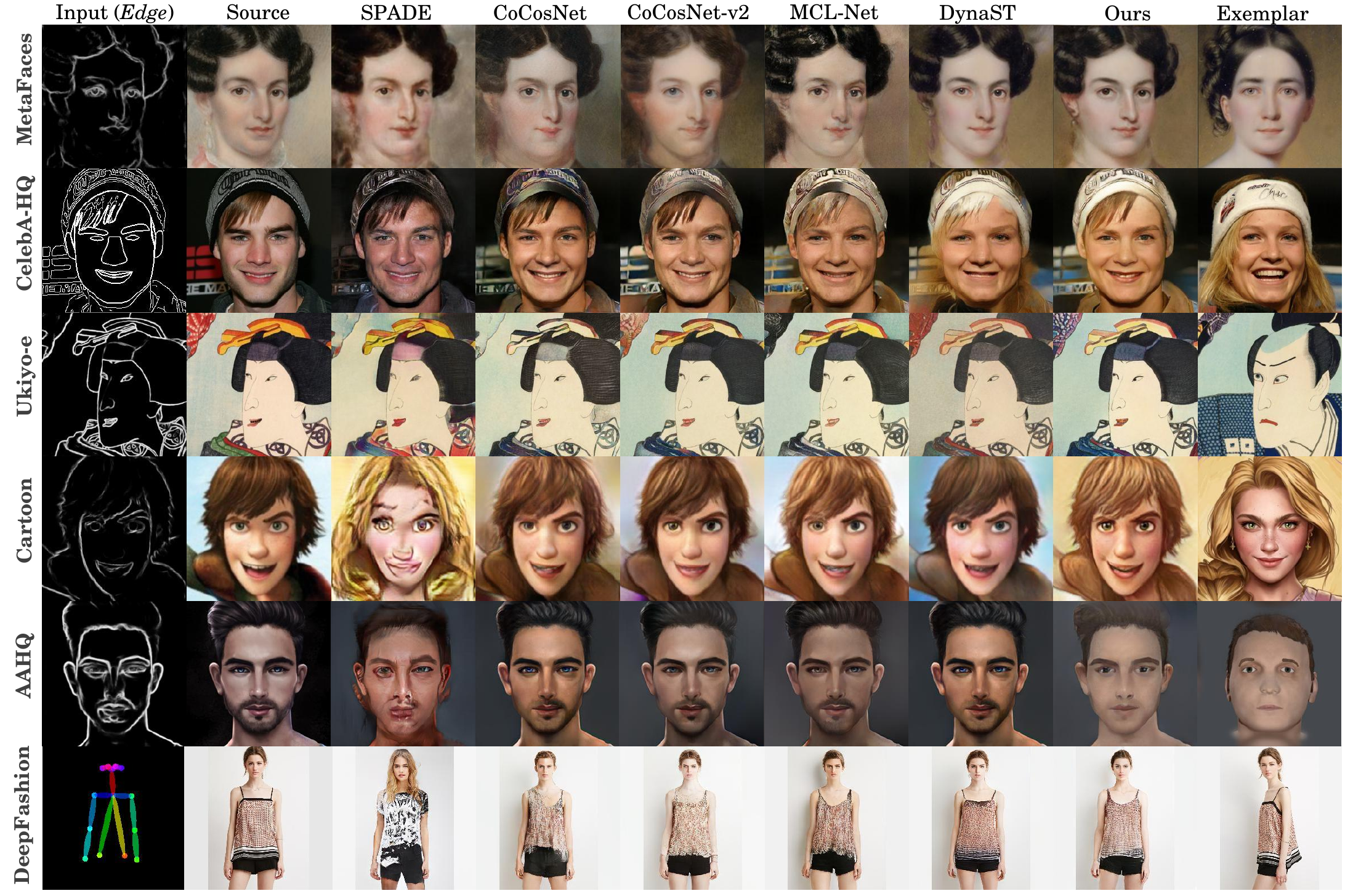}
	 \vspace{-0.7cm}
	\caption{Results on the Metfaces \cite{karras2020training}, CelebA-HQ \cite{lee2020maskgan}, Ukiyo-e \cite{pinkney2020ukiyoe}, Cartoon \cite{pinkney2020ukiyoe}, AAHQ \cite{liu2021blendgan}, and DeepFashion \cite{DeepFashion} datasets. 
	}
	\label{fig:comp}
	 \vspace{-0.6cm}
\end{figure*}

\section{Experiments}
\label{sec:exp}

\textbf{Implementation details.} 
We apply spectral normalization \cite{miyato2018spectral} to all the layers in the translation network and discriminator. We use the Adam \cite{kingma2014adam} solver with $ \beta_{1}  = 0$ and $ \beta_{2}  = 0.999$. The learning rates for the generator and discriminator are set as $1e-4$ and $4e-4$ respectively, following TTUR \cite{heusel2017gans}. The experiments are conducted using 4 24GB RTX3090 GPUs. Limited by the computation load, we restrict the resolution of generated images to $256 \times 256$ in all translation tasks. 

\textbf{Datasets.} We mainly conduct experiments on the following datasets. 
(1) \textbf{CelebA-HQ} \cite{lee2020maskgan} contains 30,000 facial photos. We chose 24,000 samples as the training set and 3000 as the test set. 
(2) \textbf{Metfaces} \cite{karras2020training} consists of 1336 high-quality artistic facial portraits. 
(3) \textbf{AAHQ} \cite{liu2021blendgan} consists of high-quality facial avatars. We randomly select 1500 samples for training and 1000 samples for testing. 
(4) \textbf{Ukiyo-e} \cite{pinkney2020ukiyoe} consists of high-quality Ukiyo-e faces. We randomly select 3000 and 1000 samples for training and testing, respectively.
(5) \textbf{Cartoon} \cite{pinkney2020ukiyoe} consists of 317 cartoon faces. 
(6) \textbf{DeepFashion} \cite{DeepFashion} consists of 800,00 fashion images.
On CelebA-HQ, we connect the face landmarks for face region, and use Canny edge detector to detect edges in the background. On DeepFashion, we use the officially provided landmarks as input. On the other datasets, we use HED \cite{xie2015holistically} to obtain semantic edges. 


\subsection{Comparison with state-of-the-art}
\label{ssec:comp}

We select several advanced models, including SPADE \cite{park2019SPADE}, CoCosNet \cite{zhang2020cocosnet}, CoCosNet-v2 \cite{zhou2021cocosnetv2}, MCL-Net \cite{zhan2022marginal}, and DynaST \cite{DynaST}, for comparison. 
For a fair comparison, we retrain their models at resolution $256 \times 256$ under the same settings as ours.

\textbf{Quantitative evaluation.} 
We adopt several criteria to fully evaluate the generation results. 
(1) \textit{Fr$\mathrm{\acute{e}}$chet Inception Score} (FID) \cite{Seitzer2020FID} and \textit{Sliced Wasserstein distance} (SWD) \cite{karras2017progressive} are used to evaluate the image perceptual quality. 
(2) To assess style relevance and semantic consistency of translated images \cite{zhang2020cocosnet}, we compute the \textit{color}, \textit{texture}, and \textit{semantic} metrics based on VGG19 \cite{simonyan2014very}. Specifically, the cosine similarities between low-level features (i.e. $relu1\_2$ and $relu2\_2$) are used to measure \textit{color} and \textit{texture} relevance, respectively; the average cosine similarity between high-level features (i.e. $relu3\_2$, $relu4\_2$, and $relu5\_2$) measures the \textit{semantic} consistency. 

The quantitative comparison results are shown in Table \ref{tab:comp}. Compared to existing methods, our model consistently achieves superior or highly competitive performance across all the datasets. Especially, MATEBIT significantly improves the style relevance in both texture and color. 
On the complicated AAHQ dataset, which contains diverse styles of avatars, MATEBIT dramatically decreases both  FID and SWD. 
Such superiority indicates that our generated images are of better perceptual quality; and present consistent appearance to similar parts in exemplars. 
We additionally report the average time each method costs for generating an image. Our method shows the best efficiency and is significantly faster than previous methods. 


\begin{figure}
	\centering
	\includegraphics[width=1\linewidth]{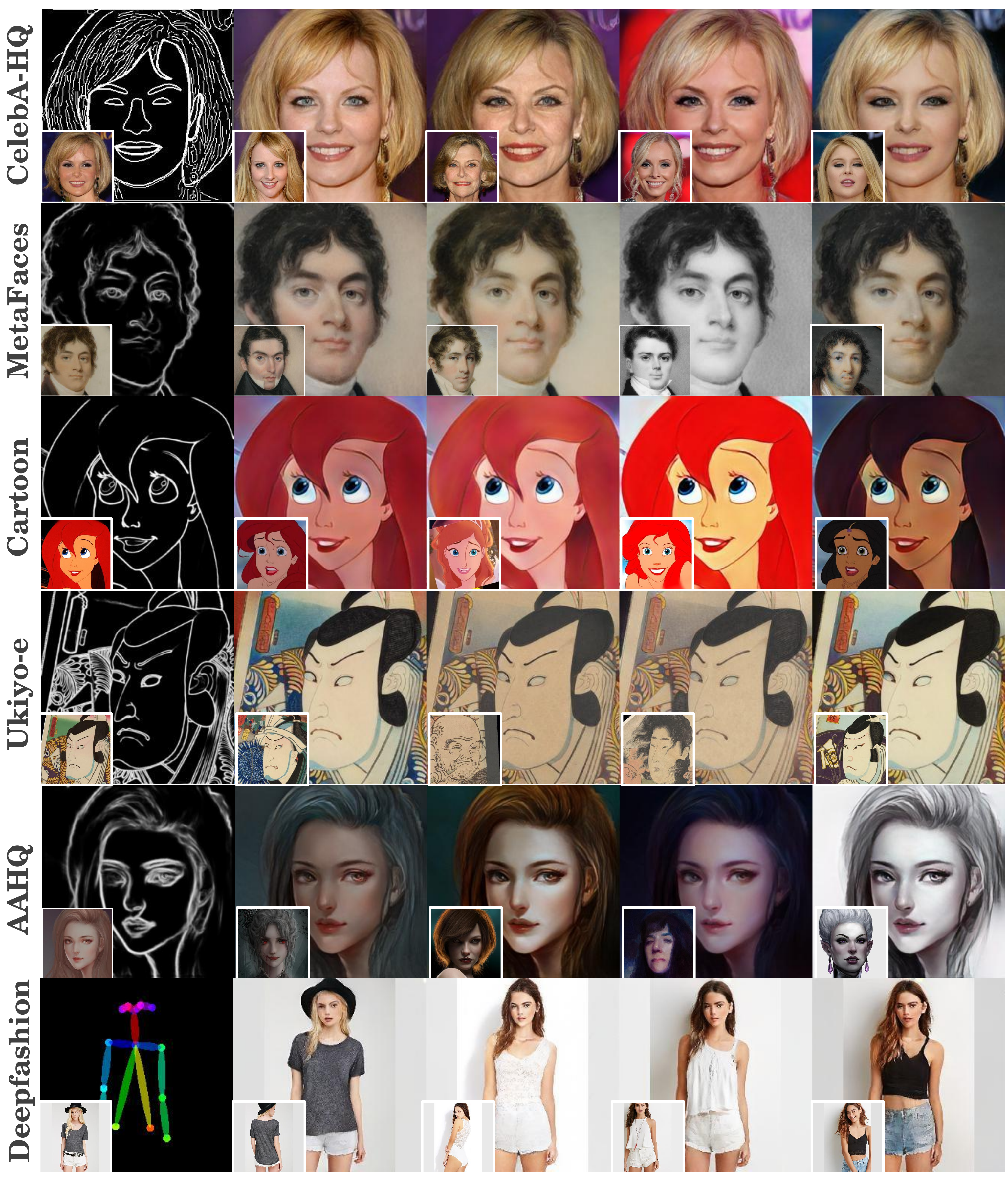}
		 \vspace{-0.6cm}
	\caption{More results generated by MATEBIT. Images shown in the left-bottom corner are source images or exemplars.}
	\label{fig:more}
	 \vspace{-0.6cm}
\end{figure}

\textbf{Qualitative comparison.} Fig \ref{fig:comp} illustrates images generated by different methods. 
Obviously, previous methods present geometric distortions, blurring artifacts, inconsistent colors, or identity inconsistency. 
In contrast, MATEBIT consistently produces appealing results, including more results shown in Fig. \ref{fig:more}. Specially, our results preserve the semantic structure of input images, and present consistent appearance with semantically similar regions in exemplars. 
Previous methods suffer serious degradations mainly due to the matching errors in full correspondence learning. In our method, we distinct reliable and unreliable correspondence, and release the role of matching in image generation. As a result, our method stably transfers a source image to the target style of an exemplar. 
\subsection{Ablation study}
\label{ssec:ablation}

\begin{figure}
	\centering
	\includegraphics[width=1\linewidth]{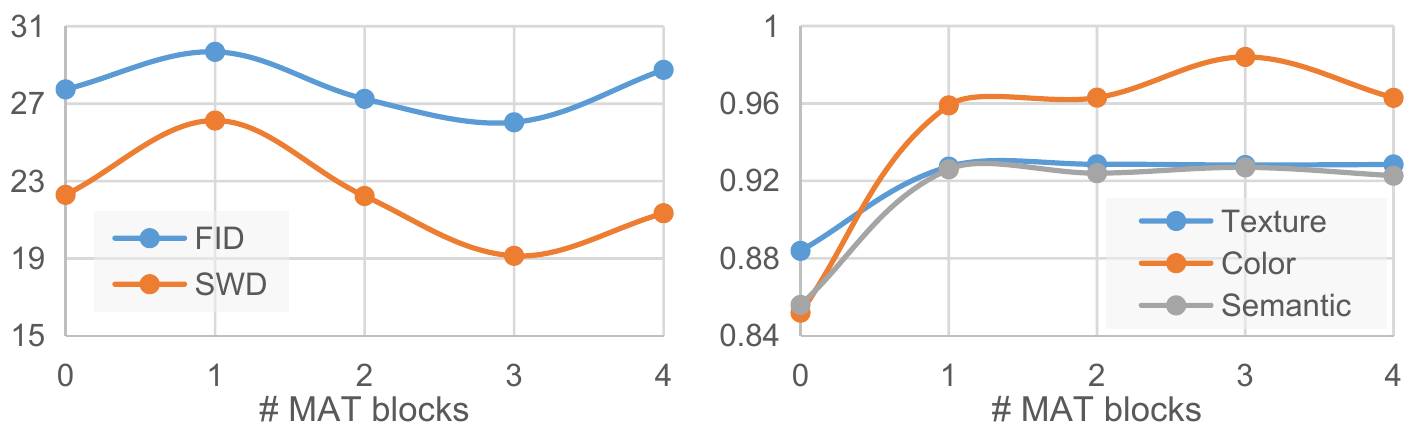}
		 \vspace{-0.6cm}
	\caption{Impact of the number of MAT blocks on performance.}
	\label{fig:nmat}
	 \vspace{-0.4cm}
\end{figure}
\begin{figure*}
	\centering
	\includegraphics[width=1\linewidth]{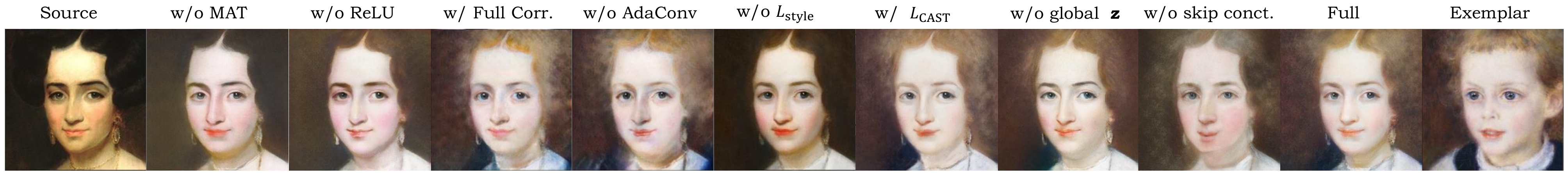}
		 \vspace{-0.6cm}
	\caption{Comparison of generated images by different variants of our model, on Metfaces \cite{karras2020training}.}
	\label{fig:ablation}
	 \vspace{-0.6cm}
\end{figure*}

\textbf{Impacts of MAT.} 
We present a comprehensive analysis to justify the important component in our architecture, i.e. MAT. We here modify our full model by (1) removing the MAT module (i.e. w/o MAT), (2) removing ReLU in MAT (i.e. w/o ReLU), (3) replacing MAT with three-layer full correspondence learning modules \cite{zhang2020cocosnet} (i.e. \textit{Full Corr.}), and (4) replacing the AdaConv with FFN \cite{zhang2020cocosnet} (i.e. \textit{w/ AdaConv}). 
The results in Table \ref{tab:ablation} show that removing MAT or ReLU dramatically hurts the performance. Besides, using the full correspondence learning in \cite{zhang2020cocosnet} or using FFN also significantly decreases the texture relevance and semantic consistency. Correspondingly, these model variants leads to inferior results in terms of textures or colors, compared to our full model (Fig. \ref{fig:ablation}). 
Recall the visualized correspondence in Fig. \ref{fig:visatt}, our method learns remarkably accurate correspondence, which ultimately benefits the quality of generated images. In addition, Fig. \ref{fig:nmat} shows that both the semantic consistency and style realism broadly improve with the number of MAT blocks and peak at three. 
All these observations demonstrate our motivation that MAT gradually refines cross-domain correspondence and augments informative representations for generating high-quality images.

\begin{table}
	\begin{center}
		\caption{Results of ablation studies on MetaFaces.}
		 \vspace{-0.3cm}
		\label{tab:ablation}
		\renewcommand\arraystretch{1.2}
		\resizebox{1\linewidth}{!}{
			\begin{tabular}{l|ccccc}
				\toprule
				& FID $\downarrow$ & SWD $\downarrow$ & Texture $\uparrow $ & Color $\uparrow $ & Semantic$\uparrow $ \\
				\midrule
				w/o MAT	&	27.7	&	22.3	&	0.883 	&	0.852 	&	0.856 	\\
				w/o ReLU &	30.1	&	20.0	&	0.916	&	0.956	&	\textbf{0.928}	\\
				w/o MAT ($\to$ Full Corr. \cite{zhang2020cocosnet})	&	34.1	&	19.7	&	0.872 	&	0.969 	&	0.902 	\\
				w/o AdaConv ($\to$ FFN \cite{vaswani2017attention})	&	34.3	&	20.1	&	0.841 	&	0.971 	&	0.896 	\\
				\midrule											
				w/o $\mathcal{L}_{style}$	&	30.3	&	20.8	&	0.874 	&	0.848 	&	0.908 	\\
				w/o $\mathcal{L}_{style}$ ($\to \mathcal{L}_{\mathrm{CAST}}$ \cite{zhang2020cast} )	&	31.0	&	\underline{19.8}	&	0.904 	&\underline{0.983 }	&	0.921 	\\
				\midrule											
				w/o $\mathcal{L}_{str}$	&	28.4 	&	21.8	&	0.915 	&	\underline{0.983 }	&	0.911 	\\
				w/o skip connection	&	47.0	&	27.7	&	0.925 	&	0.958 	&	0.902 	\\
				w/o global style $\mathbf{z}$	&	\underline{27.6}	&	21.2 	&	\underline{0.927} 	&	0.961 	& 0.925 	\\
				\midrule											
				Full Model	&	\textbf{26.0}	&	\textbf{19.1} &	\textbf{0.938} &\textbf{0.984} & \underline{0.927}	\\
				\bottomrule
			\end{tabular}
		}
	\end{center}
	 \vspace{-0.9cm}
\end{table}

\textbf{Contrastive Style Loss.} 
To verify the effectiveness of the proposed contrastive style learning methodology, we train our model by (1) removing the style loss (i.e. w/o $\mathcal{L}_{style}$) and (2) replacing $\mathcal{L}_{style}$ with the loss used in CAST \cite{zhang2020cast} (i.e. w/ $\mathcal{L}_{\mathrm{CAST}}$). 
In CAST, only high-quality exemplars in different styles are used as negative samples. Differently, we use low-quality generated images in diverse styles as negative samples. 
From both Table \ref{tab:ablation} and Fig. \ref{fig:ablation}, we observe that: (1) without $\mathcal{L}_{style}$, although the generated images show high semantic consistency, they present low style relevance; (2) $\mathcal{L}_{\mathrm{CAST}}$ benefits the style relevance, but leads to inferior performance to $\mathcal{L}_{style}$. These comparison results meet our expectation that: our CSL methodology enables the learned style codes to discriminate subtle divergences between images with different perceptual qualities. Such discriminability facilitates pushes the network to generate high-quality images.



\textbf{Skip connections \& global style control.} In MATEBIT, we use skip connections to supplement input semantic information. Removing skip connections dramatically hurts the semantic inconsistency and the quantitative results. Besides, using global style vector $\mathbf{z}$ increases subtle details, e.g. the colors over the mouth, rings, and hairs. 

In summary, MAT learns accurate correspondence and enables context-aware feature augmentation; the contrastive style learning benefits the style control and high-quality image generation; and the U-Net architecture helps the preservation of semantic information. Ultimately, all these benefits make our model significantly outperform previous state-of-the-art methods in generating plausible images. 


\begin{figure}
	\centering
	\includegraphics[width=1\linewidth]{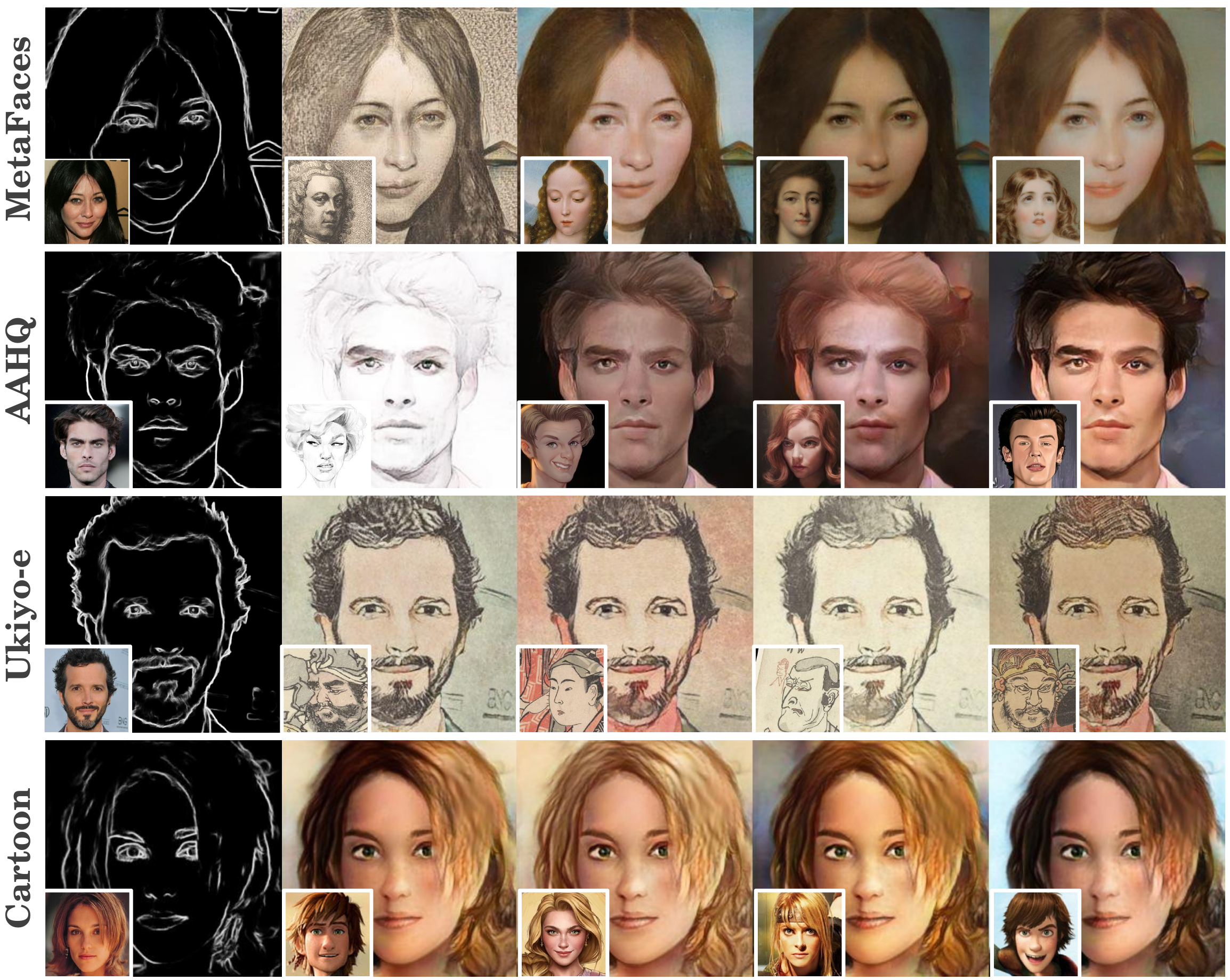}
	 \vspace{-0.6cm}
	\caption{Our method can transfer facial photos to artistic portraits in the style of exemplars. }
	\label{fig:apg}
	 \vspace{-0.3cm}
\end{figure}

\subsection{Applications} 
\label{ssec:app}


\textbf{Artistic Portrait Generation.}
An potential application of our method is transferring a facial photo to an artistic portrait, in the style of an exemplar. We here apply the previously learned models to randomly selected facial photos from CelebA-HQ \cite{lee2020maskgan}. As illustrated in Fig. \ref{fig:apg}, our method can generate appealing portraits with consistent identity and faithful style appearance.

\begin{figure}[t]
	\centering
	\includegraphics[width=0.97\linewidth]{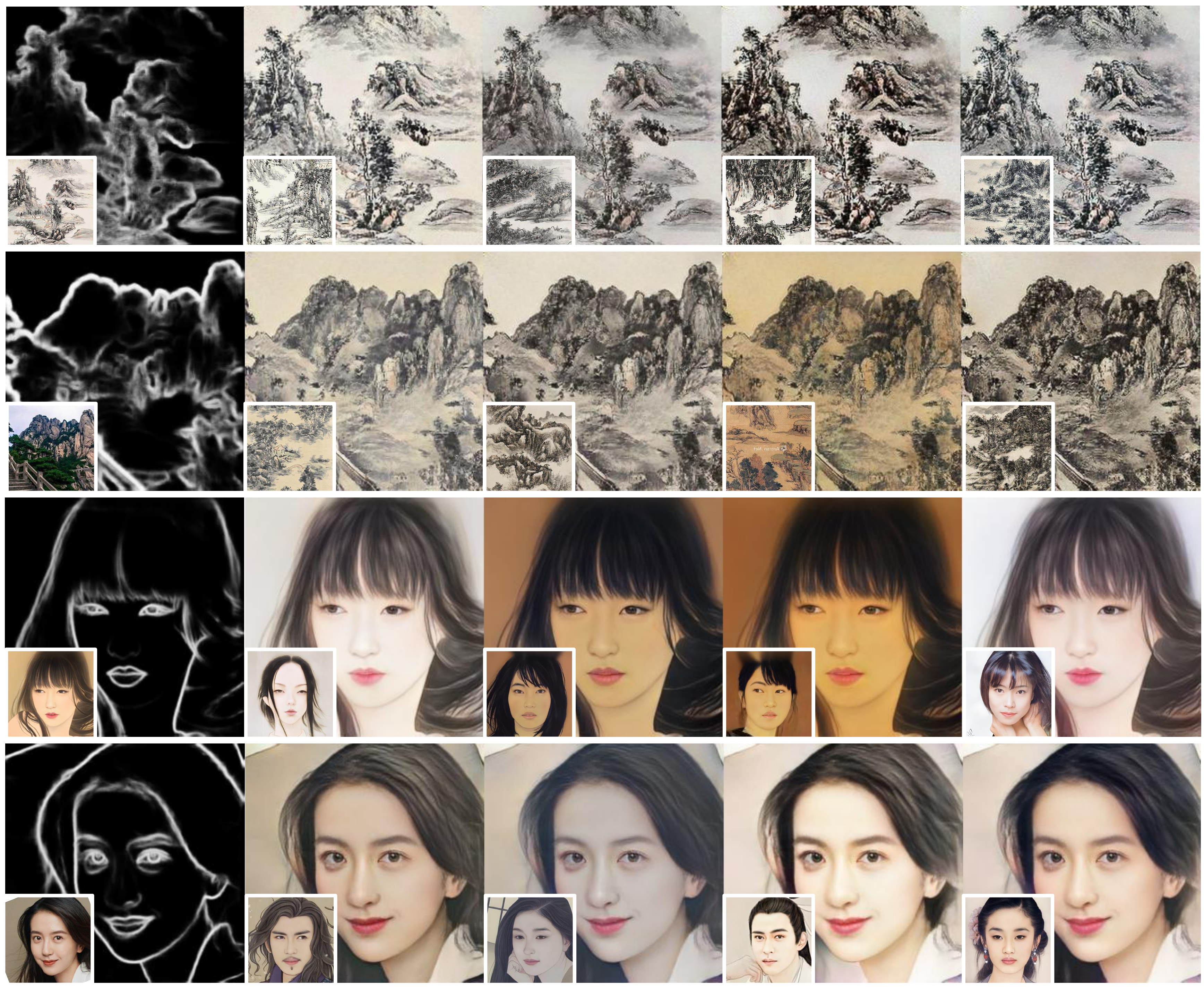}
	 \vspace{-0.2cm}
	\caption{Chinese ink paintings generation (1st \& 3rd rows), as well as photo-to-painting translation (2nd \& 4th rows). 
	}
	\label{fig:land}
	 \vspace{-0.6cm}
\end{figure}

\textbf{Chinese Ink Painting Generation.} To verify the capacity of our model in generating complex images, we additionally apply it to generate Chinese Ink paintings. Specially, we collect paintings of landscapes and facial portraits from the web, and then train and test our model on each subset respectively. Fig. \ref{fig:land} illustrates the results of painting generation and photo-to-painting translation. Obviously, all the generated images show remarkably high quality. Besides, our model successfully captures subtle differences between different exemplars, demonstrating its remarkable capacity in style control.

\section{Conclusions}
\label{sec:conlusion}
This paper presents a novel exemplar-guided image translation method, dubbed MATEBIT. Both quantitative and qualitative experiments show that MATEBIT is capable of generating high-fidelity images in a number of tasks. Besides, ablation studies demonstrate the effectiveness of MAT and contrastive style learning.  
Despite such achievements, the artistic portraits transferred from facial photos (Fig. \ref{fig:apg}) are inferior to those shown in Fig. \ref{fig:more}. This may be due to the subtle differences in edge maps between photos and artistic paintings. In the near future, we will explore to solve this issue via semi-supervised learning or domain transfer technologies. 


{\small
\bibliographystyle{ieee_fullname}
\bibliography{egbib}
}

\end{document}